\documentclass[journal,twocolumn]{IEEEtran}
\usepackage[utf8]{inputenc}

\usepackage{amsmath,amssymb}

\usepackage{biblatex}
\usepackage{graphicx}
\usepackage{subcaption}
\usepackage{url}
\usepackage{wrapfig}

\newcommand{\vecX}{\mathbf{x}}

\begin{document}

\title{Recurrent Convolutional Neural Networks help to predict location of Earthquakes}
\author{Roman Kail, IITP RAS, \\
Alexey Zaytsev, Evgeny Burnaev, Skoltech}


\maketitle

\begin{abstract}  
We examine the applicability of modern neural network architectures to the midterm prediction of earthquakes.
Our data-based classification model aims to predict if an earthquake with the magnitude above a threshold takes place at a given area of size $10 \times 10$ kilometers in $10$-$60$ days from a given moment.
Our deep neural network model has a recurrent part (LSTM) that accounts for time dependencies between earthquakes and a convolutional part that accounts for spatial dependencies.
Obtained results show that neural networks-based models beat baseline feature-based models that also account for spatio-temporal dependencies between different earthquakes. 
For historical data on Japan earthquakes our model predicts occurrence of an earthquake in $10$ to $60$ days from a given moment with magnitude $M_c > 5$ with quality metrics ROC AUC $0.975$ and PR AUC $0.0890$, making $1.18 \cdot 10^3$ correct predictions, while missing $2.09 \cdot 10^3$ earthquakes and making $192 \cdot 10^3$ false alarms.
The baseline approach has similar ROC AUC $0.992$, number of correct predictions $1.19 \cdot 10^3$, and missing $2.07 \cdot 10^3$ earthquakes, but significantly worse PR AUC $0.00911$, and number of false alarms $1004 \cdot 10^3$. 
\end{abstract}

\section{Introduction}

The earthquake prediction is a substantial but challenging problem~\cite{cheong2014short}.
The goal is to predict the time and location of a future earthquake.
The two common ways to solve this problem are physical modeling and machine learning based on data on past observations.

While physical modeling is a well-established approach, it covers only part of the full picture due to high uncertainties in data and complex nonlinear behavior of seismicity.
An alternative is to construct a data-based machine learning model to replace expensive and sometimes imprecise physical modeling. 
The machine learning modeling appears to work in various areas including drilling~\cite{klyuchnikov2019data}, 
high energy physics engineering~\cite{baranov2017iop},
and earthquake signal detection~\cite{mousavi2019cred}.
The machine learning serves as an alternative or a complement to physical modeling in earthquake prediction too~\cite{mignan2019deeper,mignan2019one}.

A typical machine learning model takes past information as input and outputs a future earthquake probability~\cite{panakkat2007neural,asim2017earthquake,asencio2016sensitivity,gitis2015adaptive,gitis2016approach}. 
Different models use different available data sources as inputs such as soil radon data \cite{zmazek2003application}
or history of past earthquakes itself.

Classic workflow of machine learning algorithm includes generation of precursors or another meaningful features from available information as model input.
In ~\cite{asim2017earthquake} authors used eight seismic indicators based on seismic characteristics. 
The authors of~\cite{pakistan} proceed similarly but identify that the set of useful seismic features for different regions can be different.
Other physics-inspired inputs are RTL features that aggregates the past seismic activity into a single index~\cite{RTL-Sobolev}. 
To use RTL features it is crucial to select good values of hyperparameters~\cite{proskura2019}.

More recent idea in machine learning community is to learn the right input features from data using representation learning based on Neural networks (NN).
For earthquake prediction usage of NN dates back to at least 1994~\cite{aminzadeh1994adaptive}.
Another work~\cite{alexandridis2013large} uses different class of NN Radial Basic Functions and  
in the authors conclude that Radial Basis Functions Neural Networks are also useful for large earthquakes prediction and 
adopt specific methods for training of neural networks to handle class imbalance in earthquake prediction problem.

\begin{figure}
    \centering
    \includegraphics[scale=0.45]{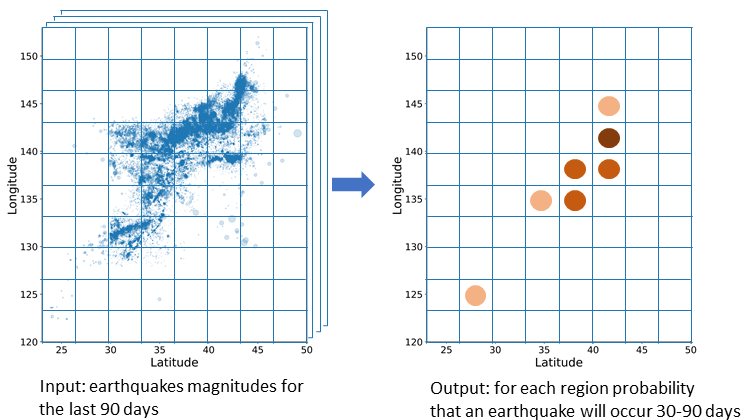}
    \caption{Workflow of our machine learning model: for each region we put a mark with an indication of an earthquake magnitude that occurs in this region; we get these indicators for the last $150$ days; the model outputs probability of an earthquake with the magnitude greater or equal to a threshold (e.g. $5$). Probabilities are indicated by color intensity in the right image.}
    \label{fig:scheme}
\end{figure}

Convolutional neural networks (CNN) also serve as machine learning models for earthquake prediction~\cite{perol2018convolutional}. 
An approach~\cite{Huang2018convolutional} uses CNNs to predict earthquakes. 
The model predicts the probability of an earthquake with a magnitude larger than $6$ happening in Taiwan in the next 30 days. 
The authors create a binary map of earthquakes of size $256 \times 256$ with each pixel is an indicator of the occurrence of an earthquake in the given region. 
These maps serve as inputs to a CNN model.
Another recent application of Recurrent Neural Networks (RNN) is in~\cite{Wang2017convolutionalLSTM}. The authors predict earthquakes taking into account temporal and spatial correlations among earthquakes. 
They predict earthquakes in $9$ sub-regions of China and use fully-connected NN block inside RNN solving the problem at a global level.


According to our knowledge nobody considers local predictions of earthquakes at small areas.
Also there is a limited usage of modern Neural Networks for earthquake prediction.

In this paper we try to cover these gaps with the following problem statement.
We divide the whole target area into sub-areas of size $\sim 10 \times 10$ km and predict earthquake occurrence in the next $10-50$ days in each sub-area separately: our prediction is local compared to the previous works.
Inputs to our machine learning model are past observations of earthquakes on the same grid, so we have a significant number of inputs and outputs.
To reduce the number of parameters, we use CNNs that able to generate useful data representations in computer vision~\cite{goodfellow2016deep} and remote sensing~\cite{ignatiev2019targeted} and learn well spatial correlations.
On top of them, we use RNNs that keep track of the past events, enabling long term memory in our model.
The general scheme for our approach is in Figure~\ref{fig:scheme}.
To train and test our model, we use a large data sample of earthquakes in Japan region.

Our main contributions are the following:
\begin{itemize}
    \item We consider middle-time-range local earthquake prediction using only information about past earthquakes.
    \item To solve this problem, we use neural networks that can handle spatial and temporal dependencies between earthquakes. Our architecture is a recurrent neural network with convolutional layers at each time step.
    The inclusion of both types of dependencies improves the model. 
    \item Our model works at a local scale and predicts the probability of earthquakes at locations of $100$ squared kilometers size. 
    \item Our model incorporates prior information on the frequency of earthquakes at a particular location within Neural network pipeline.
\end{itemize}

\section{Problem statement}

We consider earthquakes records over 26 years. 
Each earthquake has four parameters: location $(x, y)$, time $t$ and magnitude $M$.
We split the whole map  into the grid of size $200 \times 250$ with each cell is $\sim 10$ km long and $\sim 10$ km wide and predict earthquakes at each cell.
Thus, if we identify an earthquake in a cell, the location is precise enough for most applications.

Our goal is to construct a model that predicts if there is an earthquake in the time cylinder $[T + T_{\min}, T + T_{\max}]$ for each cell using earthquake historical information up to time $T$.
We consider middle time range earthquake prediction with $T_{\min} = 10$ and $T_{\max} = 50$ days.
To identify limits of applicability of our model we consider different thresholds $M_{c} = 3.5$ and $M_{c} = 5$ for the earthquake magnitude: our target is to predict earthquakes with $M \geq M_c$.
While smaller thresholds $M_c$ are better from the machine learning point of view, as the class imbalance is less severe, higher values of $M_c$ are more interesting from a practical point of view.

We have pairs $(\vecX_i, y_i)$, where $y_i$ is the indicator of an earthquake with a magnitude $M \geq M_c$, and $\vecX_i$ is the vector of features that represents times and specific locations of past earthquakes.
All these pairs form a sample $D = \{(\vecX_i, y_i)\}_{i = 1}^n$.
Our goal is to create a model $\hat{y}(\vecX)$ that is as close as possible to the true value $y(\vecX)$.
Thus, the considered problem is a machine learning classification problem.

\subsection{Data}

\begin{figure}
    \centering
    \includegraphics[scale=0.3]{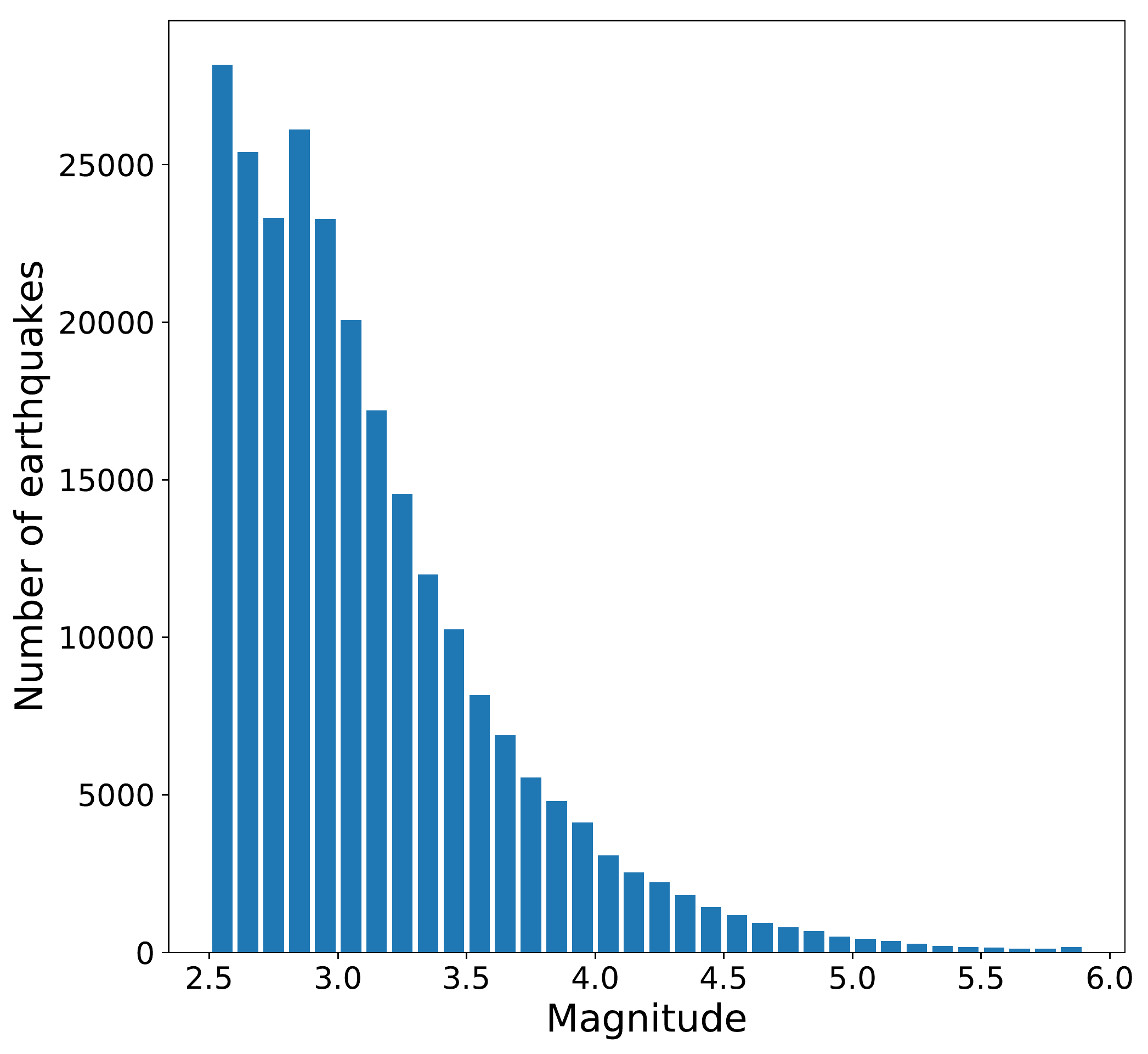}
    \caption{Histogram of magnitudes: the total sample size is about $250 000$. Only earthquakes with magnitude smaller or equal to $6$ are presented. There are only $350$ earthquakes with a magnitude greater or equal to $6$ and only $2387$ earthquakes with magnitude greater or equal to $5$. Also we see a local maximum at $3$, as our grid of observational stations are not dense enough, and we miss some earthquakes with smaller magnitudes.}
    \label{fig:hist_magnitudes}
\end{figure}

In our work we consider a dataset from a typical seismic-active region - Japan. 
The dataset consists of data about $247 204$ earthquakes that occurred
in $1990-2016$.
Figure~\ref{fig:hist_magnitudes} demonstrates the histogram of the number of earthquakes with respect to the magnitude. 

The sample is unbalanced. Most of classifiers and their accuracy metrics are tailored to balanced samples. In our case we should tune a classifier to make it more sensitive to the target class.
Also the sample is non-homogeneous as the network of seismic stations changes over time and is nonuniform. 
So we should take these into account during feature generation. 



\section{Methods}
\label{sec:methods}
\subsection{Naive baseline}
\label{naive}

A rather strong baseline for prediction of an earthquake probability at a given location is usage of a historical mean occurrence value at this subarea. 

\subsection{RTL features}
\label{RTL_features}

Another approach for the midterm prediction of earthquakes is usage of RTL features (Region-Time-Length) to identify the probability of an earthquake in a given region~\cite{RTL-Sobolev}. 
RTL is composed of weighted quantities associated with three parameters: time, location and magnitude of earthquakes and depend on past earthquakes in space-time cylinder. 
So, the model based on RTL features as inputs is an adequate baseline as it takes into account these spatial and temporal dependencies between earthquake events.



\subsection{Ensembles of decision trees}

Typical nonlinear baseline in classification problems is an Ensemble of basic decision trees classifiers.
The advantages of this approach include a decent performance with default
settings~\cite{fernandez2014we}, fast model construction, almost no
over-fitting and handling of various problems in data including missing values and outliers.

Among various approaches for construction of Ensembles of Decision Trees,
the most used nowadays is Gradient Boosting~\cite{friedman2001greedy}.
Modern implementations serve as baseline in many practical problems~\cite{chen2015xgboost} and suitable for imbalanced classification problems~\cite{kozlovskaia2017deep}.


As inputs to Gradient Boosting classifiers we use either RTL features with a set of hyperparameters $r_0, t_0$ equal to the optimal ones from~\cite{proskura2019} or binary indicators of earthquakes for previous days.

\subsection{Neural networks}
\label{our_approach}
\subsubsection{RNNs}
We adopt recurrent neural networks (RNNS) to capture temporal dependencies. In particular, we use LSTM, a type of RNN which uses additional state cell to enable long-term memory~\cite{gers1999learning}.
In our case, we use two-dimensional feature maps as hidden states compared to one-dimensional hidden states in common LSTM, as we need some map-to-map transformations.

\subsubsection{CNNs}
We process a distribution of earthquakes on the map of $200 \times 250$ cells. To benefit from spatial dependencies between earthquakes we use convolution neural networks~\cite{lecun1998gradient}, which efficiently work with images or other two-dimensional signals like remote-sensing data.




Our approach combines RNN and CNN architectures, as we pass information through RNN in a form of a feature map, obtained using CNN, see paragraph \ref{subsec:full}.

\subsection{Residuals normalization for neural networks}
\label{subsec:residuals}

The naive classifier~\ref{naive} performs quite well. 
To take advantage of it we adjust predicted earthquake probabilities with naive classifier outputs as priors at the last neural network layer.
We make an inverse softmax transformation of prior predictions at each cell. Then we add the neural network output and combine NN and prior output, applying common softmax transformation in the end.
The following procedure for classes $i \in \{1, 2\}$ is used:
\begin{itemize}
    \item An output of our model is $\delta o_i$. The naive prediction before softmax equals $o_i = \log p_i + c$, where $p_i$ is the earthquake prior probability at a given location, $c$ is a hyperparameter that scales the power of this prior.
    \item We calculate $o_i + \delta o_i$.
    \item The probability of an earthquake as the prediction of the neural network is $\hat{y}_i = \mathrm{softmax}(o_i + \delta o_i)$.
    \item We use a threshold (e.g. $t = 0.5$): if $\hat{y}_2 > t$, the model reports an earthquake.
\end{itemize}
During training we use the cross-entropy loss function for pairs of predicted probabilities $\hat{y}_i$ and true labels $y_i$.

\subsection{Full pipeline}
\label{subsec:full}

\begin{figure}[h]
    \centering
    \includegraphics[scale=0.25]{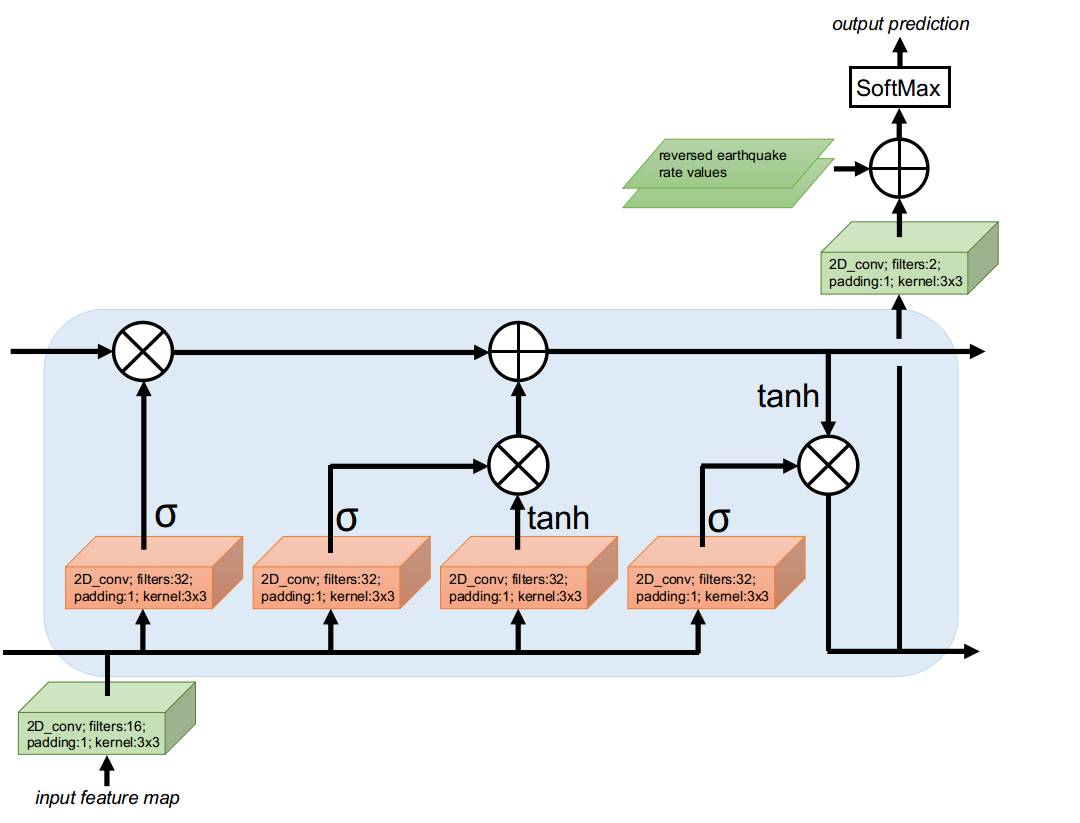}
    \caption{Proposed CNN-LSTM model. Green blocks represent series of convolutions, orange blocks represent convolution layers, circles --- multiplications and additions, and $\sigma$ and $\tanh$ --- sigmoid and hyperbolic tangent activation functions correspondingly.}
    \label{fig:network_architecture}
\end{figure}

Our main architecture depicted in Figure~\ref{fig:network_architecture} follows the pipeline:
\begin{enumerate}
  \item We represent data as a sequence of heat maps: for each cell we specify a magnitude of an earthquake on this day; we set it to zero if no earthquake happened. The input heat map at each time moment has size $200 \times 250$.
  \item We pass the input heat map through a convolutional network to create an embedding of size $200 \times 250$ with $16$ channels. As an output of LSTM at each time moment we have a hidden representation (short term memory) of size $32 \times 200 \times 250$, cell (long term memory) representation of a similar size, and the output of size $32 \times 200 \times 250$.
  \item We transform to the output to the size $2 \times 200 \times 250$ using a series of convolutions.
  \item We follow the procedure described in paragraph~\ref{subsec:residuals} to modify our prediction according to the prior probabilities of earthquakes at given locations.
\end{enumerate}

We also tried U-net architecture~\cite{ronneberger2015u}, often used for image segmentation, instead of the convolutional architecture, but results were~worse.

\section{Results}

\begin{table}[h]
    \centering
    \begin{tabular}{lcccc}
    \hline
    Weight & \multicolumn{2}{c}{$M_c = 3.5$} & \multicolumn{2}{c}{$M_c = 5$} \\
    & ROC AUC & PR AUC & ROC AUC & PR AUC \\
    \hline
    $1$      & 0.643 & 0.0323 & 0.517 & 0.00044 \\
    $10$     & 0.909 & 0.0345 & 0.493 & 0.00050 \\
    $10^3$   & 0.952 & {\bf 0.0806} & 0.890 & 0.00096 \\
    $10^5$   & 0.948 & 0.0761 & {\bf 0.935} & {\bf 0.00133} \\
    $10^7$   & {\bf 0.961} & 0.0797 & 0.911 & 0.00117 \\
    \hline     
    \end{tabular}
    \caption{Dependence of CNN+LSTM model quality on the weights of the minor class objects during training. Optimal weights are significantly better than $1$, while the usage of too large weights for $M_c = 5$ leads to performance degradation.}
    \label{tab:weight_seleciton}
\end{table}

\begin{table}[h]
    \centering
    \begin{tabular}{lcccc}
    \hline
    Method & \multicolumn{2}{c}{$M_c = 3.5$} & \multicolumn{2}{c}{$M_c = 5$} \\
    & ROC AUC & PR AUC & ROC AUC & PR AUC \\
    \hline
    Baseline                         & 0.901 & 0.052 & 0.674 & 0.00198 \\
    GradBoost, indicator      & 0.578 & 0.0573 & 0.549  & 0.00021 \\
    GradBoost, RTL            & 0.754 & 0.0139  & 0.820  & 0.00088 \\
    CNN                                    & 0.950 & 0.0179 & 0.808 & 0.00044 \\
    CNN+LSTM                             & 0.952 & 0.0806 & 0.937 & 0.00152 \\
    CNN, resid.                         & 0.966 & 0.0166 & \textbf{0.994} & 0.00776 \\ 
    CNN+LSTM, resid.       & \textbf{0.975} & \textbf{0.0890} & 0.992  & \textbf{0.00911} \\ 
    \hline     
    \end{tabular}
    \caption{Dependence of the earthquake model quality on the used method. Our method outperforms other options; also usage of both CNN and LSTM along with residuals training provides further improvement. At the same time Gradient Boosting performs worse than the baseline.}
    \label{tab:methods_comparison}
\end{table}

\begin{table}[h]
    \centering
    \begin{tabular}{cccccc}
    \hline
    \multicolumn{3}{c}{Our approach (CNN+LSTM, resid.)} & \multicolumn{3}{c}{Baseline} \\
    Threshold & FN & FP & Threshold & FN & FP\\
    \hline
    \multicolumn{6}{c}{$M_c = 3.5$} \\
    \hline
    0.0001  & $6.4$  & $4400$ & 0.0001  & $10.85$  & $6971$ \\
    0.1     & $16.2$ & $2120$ & 0.1     & $36.55$ & $1006$  \\
    0.99    & $78.8$ & $80$ & 0.99    & $77.20$ & $21.85$\\
    
    \hline
    \multicolumn{6}{c}{$M_c = 5$} \\
    \hline
    0.0001  & $0.81$ & $502$ & 0.0001  & $2.07$ & $1004$ \\
    0.1     & $2.09$ & $192$ & 0.3     & $2.51$ & $164.2$\\
    0.99    & $2.74$ & $92$ & 0.9    & $3.05$ & $29.79$  \\
    \hline     
    \end{tabular}
    \caption{Values in thousands of True positive (TP, detected events), False Negative (FN, undetected events), False Positive (False Alarms about future earthquakes), True Negative (TN, correct no alarm about future earthquakes) for different values of probability threshold $t$ for the best method CNN + LSTM, $M_c = 3.5$ and $M_c = 5$.}
    \label{table:confusion_matrix}
\end{table}

In this section we provide results of computation experiments. After necessary definitions we observe selection of hyperparameters for our algorithms in subsection~\ref{sec:imbalance_results} and 
then we compare all approaches in subsection~\ref{sec:comparison_of_models}.
The code for the conducted experiments is available at github~\footnote{\url{https://github.com/romakail/Earthquake_prediction_DNN}}.

\subsection{Quality metrics}

Accuracy metric is unrepresentative due to class imbalance: a constant prediction ``no-earthquake'' has a very high accuracy. 
Instead we calculate two types of errors:
number of False Negatives (FN) --- objects of the first class attributed by the classification to the second class, number of False Positives (FP) --- objects of the second class attributed by the classification to the first class.


We also use standard in machine learning community metrics ROC AUC and PR AUC.
Both these metrics lie in the interval $[0, 1]$ and their higher values  correspond to better models.
PR AUC suits better for measuring quality in imbalanced classification problems~\cite{burnaev2015influence, kozlovskaia2017deep}.

\subsection{Compared algorithms}

We compare the methods from Section~\ref{sec:methods}:
\begin{itemize}
    \item Baseline --- outputs mean earthquake probability at a given location obtained from historical data.
    \item Grad. boosting, indicator --- Gradient boosting with earthquake indicator input features.
    \item Grad. boosting, RTL --- Gradient boosting with earthquake RTL input features, see paragraph~\ref{RTL_features}.
    \item CNN --- series of stacked one by one convolution layers, trained on given amount of previous days.
    \item CNN+LSTM --- Recurrent neural network, which passes data as a feature map, obtained by CNN.
    \item CNN, resid. (residuals) --- same as CNN, but it predicts a residual to $o_i$, see paragraph~\ref{subsec:residuals}.
    \item CNN+LSTM, resid. (residuals) --- same as CNN+LSTM, but it predicts a residual to $o_i$, see paragraph~\ref{subsec:residuals}.
\end{itemize}


\subsection{Fighting class imbalance}
\label{sec:imbalance_results}
To deal with the class imbalance we use an oversampling technique 
increasing weights for the less populated minor class objects during training, while keeping weights for the major class objects equal to  $1$.
The dependence of model quality on the weight of the minor class objects is in Table~\ref{tab:weight_seleciton}.
In other experiments we use weight $1000$ providing performance close to optimal.

\subsection{Comparison of proposed models}
\label{sec:comparison_of_models}

In Table~\ref{tab:methods_comparison} we compare our methods based on neural networks with general Gradient boosting approach and Naive baseline. 
Both PR AUC and ROC AUC scores suggest that the performance of our models is better.
Moreover, taking into account residuals during the prediction further improves the model.
The improvement is significant for both magnitude thresholds $M_c = 3.5$ and $M_c = 5$.

Number of errors of different kind for our best classifier for a varying threshold is in Table~\ref{table:confusion_matrix}.
We see, that we can select a trade-off between the number of False alarms (FP) and the number of missed earthquakes given by our model.

\section{Conclusions}
Modern neural network architectures are good for the midterm prediction of earthquakes.
Our machine learning model predicts if an earthquake with a magnitude above a given threshold takes place at a given location in a time range of $10$-$60$ days from a selected moment.

Recurrent and Deep Convolutional Neural Networks account for time and spatial dependencies correspondingly.
A machine learning model based on these architectures provides a decent quality.
Thus, we replace hand-crafted features by features automatically extracted by neural networks, avoiding manual feature generation.

For historical data on Japan earthquakes, our model has ROC AUC $0.975$ and PR AUC $0.0890$ compared to lower values $0.992$ and $0.00911$ for the baseline approach. More intuitive quality metrics are amount of undetected events of occurence of earthquakes with magnitude $M_c > 5$  in the next $10-50$ days ($2.09 \cdot 10^3$ vs $2.07 \cdot 10^3$) and corresponding number of false alarms ($192 \cdot 10^3$ vs $1004 \cdot 10^3$) with the same amount of properly detected earthquakes on the test dataset. The proposed model significantly decreases number of false alarms and increases PR AUC.

\section{Acknowledgments}

We thank Polina Proscura, Valery Gitis and Alexander Derendyaev for useful discussions and interesting suggestions during the preparation of this work.

\printbibliography

\end{document}